\documentclass[11pt,a4paper]{article}
\usepackage[hyperref]{emnlp2021}
\usepackage{times}
\usepackage{latexsym}
\usepackage{url}
\usepackage{booktabs}       
\usepackage{amsfonts}       
\usepackage{nicefrac}       
\usepackage{microtype}      
\usepackage{algorithmic}
\usepackage{algorithm}
\usepackage{wrapfig}
\usepackage{lipsum}
\usepackage{xspace}
\usepackage{amsmath}
\usepackage{bm}
\usepackage{graphicx} 
\usepackage{subfigure} 
\usepackage[skip=5pt]{caption}
\usepackage{paralist}
\usepackage{float}
\usepackage{tabularx}
\usepackage{multirow}
\usepackage{diagbox}
\usepackage{tikz}
\usepackage{booktabs}
\usepackage{array, makecell} 
\usepackage{natbib}
\usepackage{amsmath,amssymb}
\usepackage[inline]{enumitem}
\usepackage{acronym}
\usepackage{bm}

\usepackage{microtype}

\newcommand{\mypar}[1]{\paragraph{#1}\vspace{-0.2cm}}

\title{Building and Evaluating Open-Domain Dialogue Corpora \\ with Clarifying Questions}

\author{Mohammad Aliannejadi\textsuperscript{1}, Julia Kiseleva\textsuperscript{2}, Aleksandr Chuklin\textsuperscript{3}, \\ \textbf{Jeffrey Dalton}\textsuperscript{4}, \textbf{Mikhail Burtsev}\textsuperscript{5} \\
\textsuperscript{1}University of Amsterdam,  \textsuperscript{2}Microsoft Research, 
\textsuperscript{3}Google Research \\
\textsuperscript{4}University of Glasgow,  
\textsuperscript{5}MIPT \& AIRI \\
\texttt{m.aliannejadi@uva.nl, julia.kiseleva@microsoft.com} \\ 
\texttt{chuklin@google.com,jeff.dalton@glasgow.ac.uk,burtcev.ms@mipt.ru}
}

\date{}

\begin{document}
\maketitle
\begin{abstract}
Enabling open-domain dialogue systems to ask clarifying questions when appropriate is an important direction for improving the quality of the system response. Namely, for cases when a user request is not specific enough for a conversation system to provide an answer right away, it is desirable to ask a clarifying question to increase the chances of retrieving a satisfying answer. To address the problem of \emph{`asking clarifying questions in open-domain dialogues'}: (1)~we collect and release a new dataset focused on open-domain single- and multi-turn conversations, (2)~we benchmark several state-of-the-art neural baselines, and (3)~we propose a pipeline consisting of offline and online steps for evaluating the quality of clarifying questions in various dialogues. These contributions are suitable as a foundation for further research.
\end{abstract}

\section{Introduction}
\label{sec:intro}

The ultimate goal of a conversational system is to assist users by returning an appropriate answer in response to their requests~\cite{kiseleva2016predicting, li2021data}. Recent progress on neural approaches to natural language processing~\cite{devlin2018bert, LiuRoberta_2019, clark2020electra}, and the availability of large amounts of conversational data have triggered a renaissance in end-to-end neural open-domain chatbots~\cite{adiwardana2020towards, roller2020recipes, zhang2019dialogpt, burtsev2017search, dalton2020proceedings}. There has been great progress on suggesting measures to evaluate what makes a conversation satisfying for users using various human evaluation techniques~\cite{li2019acute, see2019makes}. Those efforts showed that suggested large pre-trained models do not always perform seamlessly~\cite{see2019massively}, and there are still several challenges needed to be solved for open-domain conversational systems~\cite{huang2020challenges}.

\begin{figure}[t]
\centering
   \includegraphics[clip, width=1.0\columnwidth]{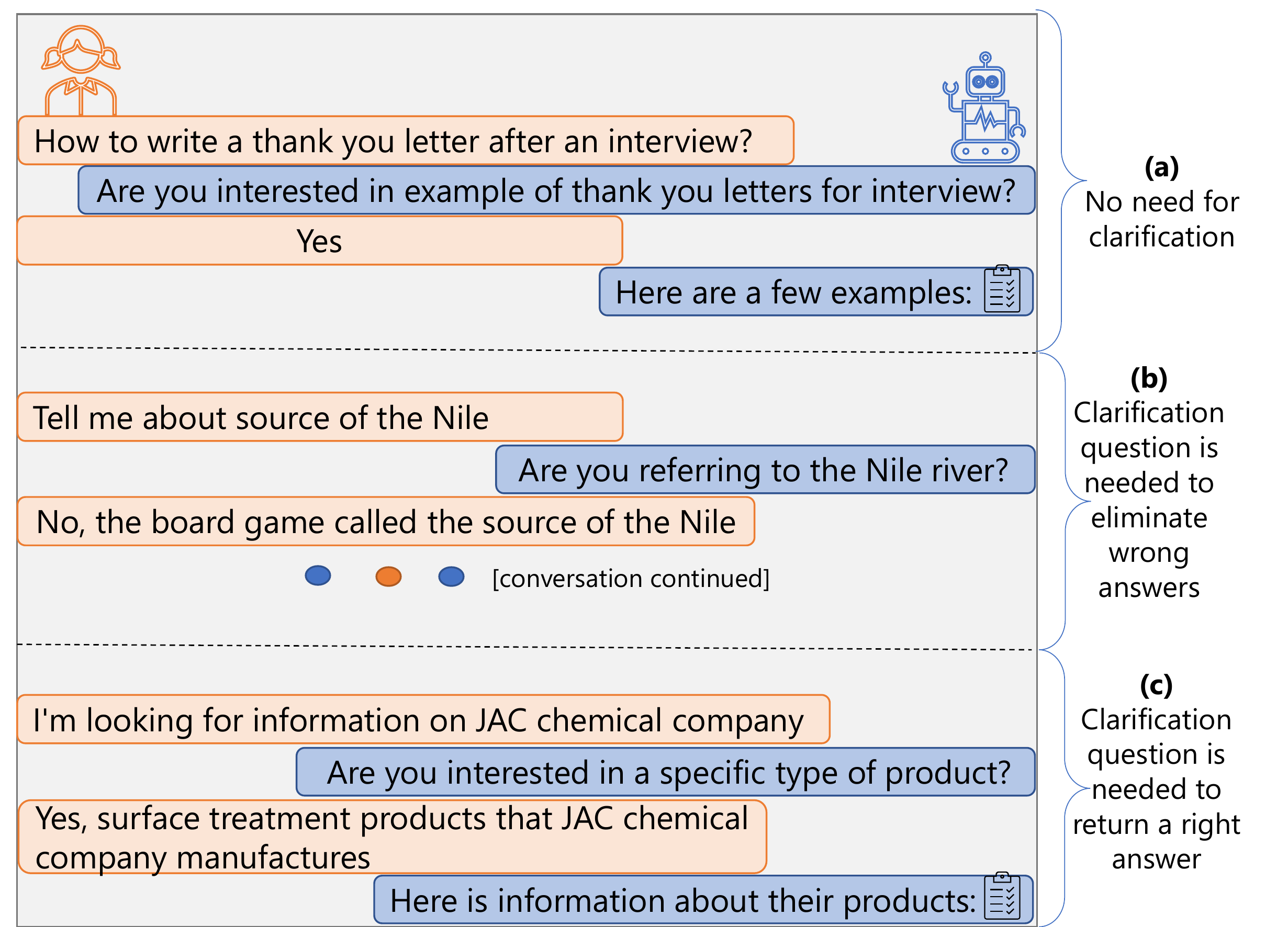}
   \caption{Examples of clarification questions embedded into the open domain conversations: (a) represents a clear user request, so clarification was unnecessary; (b) and (c) demonstrate a situation when the request is ambiguous and a system needs to act.}
   \label{fig:main_intro}
\end{figure} 

\citet{nass2000machines} conclude that people have similar expectations from talking to bots and humans. This similarity is a possible explanation for why sometimes user requests might be ambiguous and incomplete, as shown in Fig.~\ref{fig:main_intro} (b) and (c). This ambiguity is especially challenging to handle in a dialogue setting, where a system is limited by returning only one answer in response to each request, unlike in web search setup where diversification of results is possible and acceptable~\cite{vallet2012personalized}. Previous research has shown that users are much more forgiving about system mistakes if they can act on them with minimal efforts spent~\cite{kocielnik2019will, kiseleva2016understanding}. Therefore it is more appropriate to ask a clarifying question in user request ambiguity rather than generating incorrect answers. There are separate attempts to explore the following related tasks: (1) identifying a moment when the question should be asked in the course of conversation~\cite{hancock2019feed}; and (2) retrieving a clarification question~\cite{rao2018learning, wang2018learning}.
In this paper, we aim to combine these related aspects and study the following problem of \emph{generating clarifying questions for open-domain conversations}: the system must identify wherever the question is ambiguous, and, if so then instead of trying to answer it directly, it should ask a good clarifying question (Fig.~\ref{fig:main_intro}). 
One possible stumbling block preventing the community from studying the problem of open-domain clarifying question generation to enhance user experience while interacting with a conversational bot~\cite{huang2020challenges} is the lack of suitable datasets, which we address in this work. 
To summarise, the main contributions of this work are:
\begin{enumerate}[leftmargin=*,label=\textbf{C\arabic*},nosep]
    \item releasing a dataset dedicated to the problem of asking a clarifying question in open-domain dialogue systems. The dataset includes single- ($\sim$15K) and multi-turn ($\sim$1.5M) conversations and covers $\sim$300 various topics and it is suited to study: (1)~\emph{when} a clarifying question should be asked given the current context of the conversation; and (2)~\emph{which} question should be asked;
    \item benchmarking several state-of-the-art (SoTA) neural models; and 
    \item building an evaluation pipeline that provides fast iteration and involves two stages: (1) offline (automatic evaluation); and (2) online (human-in-a-loop to converse with the system).\footnote{The pipeline was designed as part of the ConvAI3~\cite{DBLP:journals/corr/abs-2009-11352} data challenge (\url{https://convai.io})}
\end{enumerate}

\noindent
We release the collected dataset, offline evaluation pipeline, and the code for running explored neural SoTA models. These models can be employed as baselines for the task.\footnote{Available at \url{https://github.com/aliannejadi/ClariQ}}

\section{Related work}
\label{sec:rel_work}
Our work is broadly relevant to two strands of research: learning to ask clarifying questions in open-domain conversational settings (Section~\ref{sec:rel_work_questions}) and evaluating dialogue systems (Section~\ref{sec:rel_work_eval}).

\subsection{Learning to ask clarifying questions}
\label{sec:rel_work_questions}
Information retrieval community has paid close attention to the problem of ambiguity in user search queries. Previously this problem was addressed through the diversification of search result pages~\cite{radlinski2006improving, kong2016precision,kong2014extending}, including via usage of personal and contextual data~\cite{jiang2015query,kato2016suggest}. Recently,~\cite{rosset2020leading, AliannejadiSigir19, zamani2020generating} suggest techniques to address ambiguity by generating clarifying questions.

Where the general settings are: (1)~a user is issuing an \emph{ambiguous} keyword query; (2)~a search engine's goal is to suggest conversational clarifying questions to help to find the required information~\cite{DBLP:conf/ictir/KrasakisAVK20,DBLP:journals/corr/abs-2103-06192,DBLP:conf/ecir/SekulicAC21, aliannejadi2021cikm}. These works also resulted in a number of datasets, e.g. \emph{Qulac}~\cite{AliannejadiSigir19} and \emph{MIMICS}~\cite{zamani2020mimics}, which consists of queries, issued by real users, and behavioral signals such as clicks. 
\citet{braslavski2017you} focus on characteristics, forms, and general patterns of clarifying questions.

Suggesting a clarifying questions is closely related to question answering (Q\&A)~\cite{kwiatkowski2019natural,DBLP:conf/eacl/SoleimaniMW21} and question generation~(QG) domains~\cite{gao-etal-2019-interconnected, chai-wan-2020-learning}. \citet{trienes2019identifying} made an attempt to understand unclear questions, \citet{li2016dialogue} suggesting an RL-based method for deciding \emph{when} to ask for user feedback in Q\&A setup.

Recently, proactive bot behavior has started to attract researchers' attention in dialogue settings yet remains rather untouched~\cite{huang2020challenges}. \citet{rao2018learning} designed a model to rank a candidate set of clarification questions by their usefulness to the given post at Stack Exchange, which targeted the problem \emph{which} question to ask. The resulted dataset was released, but it covers specific narrow topics. In contrast, \citet{hancock2019feed} focused on \emph{when} to ask a question in order to self-retrain a bot, which has been resulted in releasing a dataset. \citet{wang2018learning} studied QG techniques in application to open-domain conversations.

\subsection{Evaluating Dialogue Systems}
\label{sec:rel_work_eval}
Dialogue systems are generally separated into two types: task-oriented and open-domain. The task-oriented ones usually have clear criteria for evaluation, e.g. turn correction ratio, inappropriate utterance ratio, proxies for accuracy, and success rate~\citep{takanobu2019guided,li2016user,Su2018D3Q,li2020guided}. Despite significant efforts to introduce automatic metrics to evaluate open-domain conversations~\cite{reiter2018structured,novikova2017we,lowe2017towards}, it remains area for exploration~\cite{li2019acute, li2018dialogue, li2021data}. To the best of our knowledge the current standard approach for evaluating open-domain dialogues requires employing human assessments via crowdsourcing platforms~\cite{zhang2018personalizing,li2019acute} or engaging volunteers to participate in research competitions~\cite{burtsev2018first, dinan2020second, burtsev2020conversational,DBLP:journals/corr/abs-2009-11352}.

Therefore, we can conclude that understanding and generating open-domain clarification questions is a major component in conversational information-seeking systems, which is still under exploration. Hence, our efforts on collecting datasets and investigating the performance of the neural SoTAs are timely and useful for future research in this area.

\section{Problem Setting}
\label{sec:problem}

\begin{figure}[t]
\centering
   \includegraphics[clip, width=1.0\columnwidth]{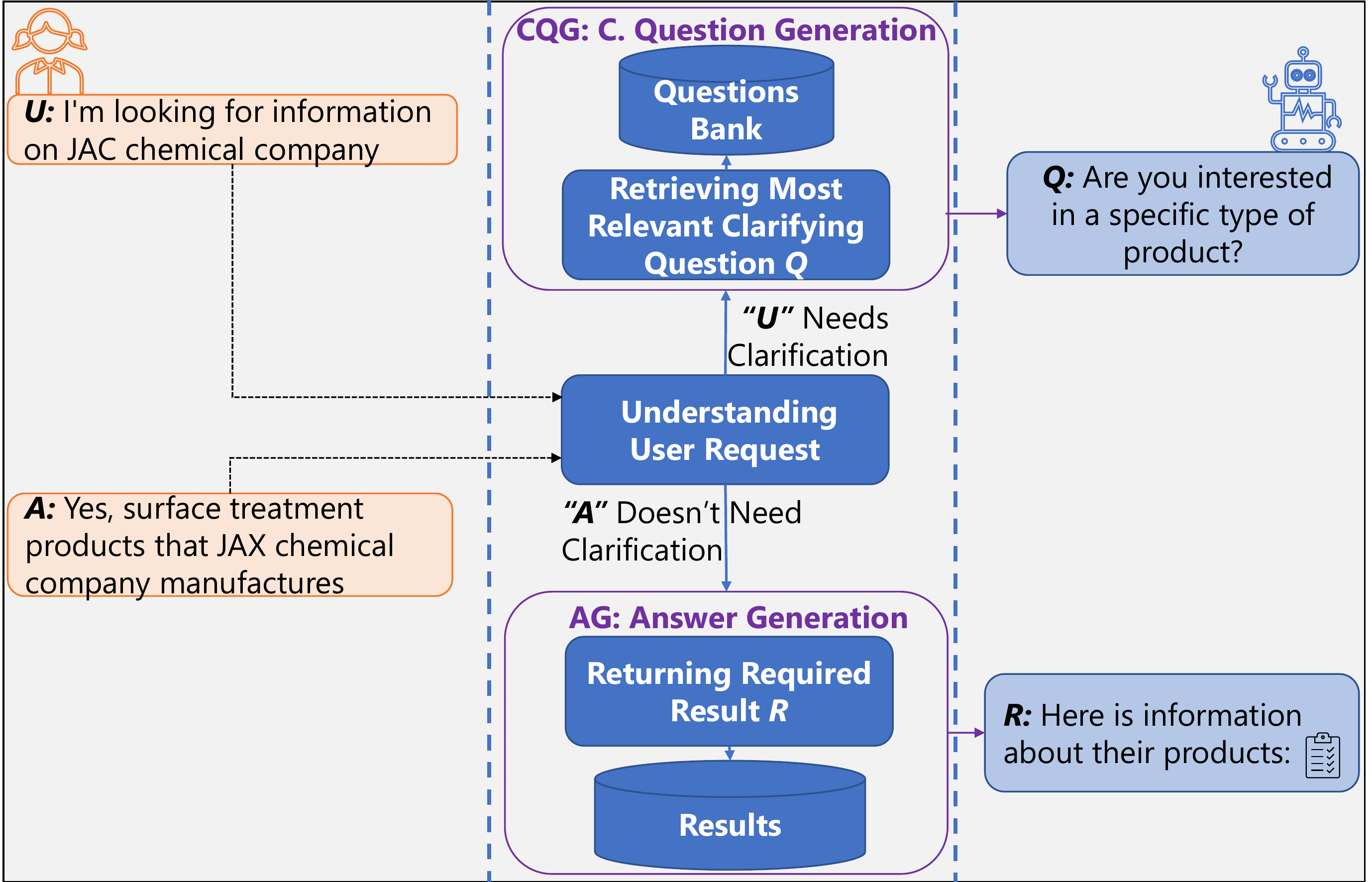}
   \caption{A pipeline for asking clarifying questions for
open-domain conversations using example Fig~\ref{fig:main_intro} (C).}
   \label{fig:problem}
\end{figure} 

Our main goal is to collect a dataset to enable studying clarifying questions generation whenever appropriate, as depicted in examples in Fig.~\ref{fig:main_intro}. Fig.~\ref{fig:problem} demonstrates a pipeline that makes it possible to process user requests in the open domain as follows: `User Request Understanding' (URU) decides which module to call either `Clarifying Question Generation' (CQG) or `Answer Generation' (AG). In this work, we focus on the first two.
We aim to collect the following data:
\begin{itemize}[leftmargin=*,nosep]
    \item \textbf{User Request ($U$):} an initial user request in the conversational form, e.g., \emph{`What is Fickle Creek Farm?'} with a label reflecting whether clarification is needed;
    \item \textbf{Set of clarification questions ($\{Q\}$):} a set of possible reasonable clarifying questions that address multiple aspects/facets of $U$, e.g., $Q_1:$ \emph{`Do you want to know the location of fickle creek farm?'}, $Q_2:$ \emph{`Would you like to know the history of fickle creek farm?'}\footnote{Candidate clarifying questions should also address out-of-collection facets.};
    \item \textbf{User Answers ($A$):} each question is supplied with a user answer, e.g., the answer to $Q_1$ is $A_1:$ \emph{`No, I want to find out where can I purchase fickle creek farm products'}, the answer to $Q_2$ is $A_2:$ \emph{`I just need general information about fickle creek'} \
\end{itemize}

The collected dataset of pairs $(U, \{Q, A\})$ can be easily transformed to a set of single-turn conversations consisting of the  coherent and consistent triples $(U,Q,A)$ as shown in the example in Fig.~\ref{fig:problem}. 
We ask items in a triple, $U$, $Q$ and $A$, satisfy the following \emph{requirements}:
\begin{enumerate}[leftmargin=*,label=\textbf{R\arabic*},nosep]
    \item \label{item:req_domain} user requests $(U)$ must cover various conversational topics to represent open-domain dialogues;
    \item \label{item:req_amb} the final collection of $U$ should contain both types: ambiguous and unambiguous;
    \item \label{item:req_conversational} each inquiry $U$ to the system should be in the conversational form;
    \item \label{item:req_amb_score} the need for clarification should be predetermined as a label for each $U$ in the collection;
    \item \label{item:req_question} each clarifying question $(Q)$ should be reasonable, coherent with $U$ and address multiple facets of every ambiguous request $U$; and
    \item \label{item:req_answer} each user answer $A$ should be consistent with the clarifying question from the system.
\end{enumerate}

\begin{figure}[t]
\centering
   \includegraphics[clip, width=1.0\columnwidth]{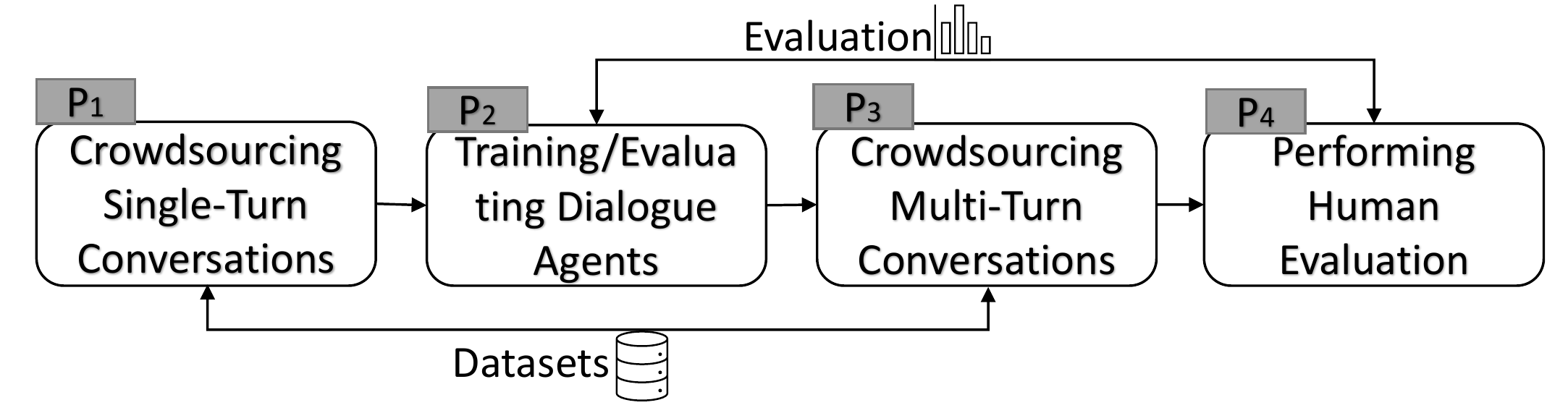}
   \caption{The pipeline highlights steps for two main goals: to collect required datasets and to perform the reproducible evaluation.}
   \label{fig:pipeline}
\end{figure}

After collecting of single-turn conversations, they are used to train various conversational agents. To collect multi-turn conversations, the two best-performing agents are utilized to converse with crowdsourced workers, who evaluate a system quality and reply to suggested clarifying questions. Finally, the two agents are evaluated using Acute-eval framework~\cite{li2019acute}, which is best available practice for online evaluation of open-domain dialogue systems.
Overall, our pipeline for \emph{data collection} and \emph{evaluation} is summarized in Fig.~\ref{fig:pipeline}.

\begin{table*}
    \caption{An example of facets for the incomplete query.}
    \label{tab:facet_example}
    \centering
    \begin{tabular}{lll}
    \toprule 
    \bf Topic  & \bf Facet        \\ 
    \midrule
    \multirow{4}{*}{Neil Young} & Find albums by Neil Young to buy\\
                                & Find biographical information about Neil Young\\
                                & Find lyrics or sheet music for Neil Young's songs\\
                                & Find a list of Neil Young tour dates.\\
    \bottomrule
    \end{tabular}
\end{table*}

\section{Data Collection} 
\label{sec:design}

Following the suggested pipeline in Fig.~\ref{fig:pipeline} this section describes our results regarding the collected datasets to initiate more follow-up studies of the problem of \emph{`asking clarifying questions in open-domain dialogues}`, namely, $P_1$: single-turn dialogues (Sec.~\ref{sec:single}) and $P_3$ multi-turn ones (Sec.~\ref{sec:multi}).

\subsection{P$_1$: Crowdsourcing Single-Turn Dialogues}
\label{sec:single}

\noindent
\mypar{\textbf{Collecting Conversational User Requests}}
Our collection of single-turn conversations is built on top of the TREC Web track 2009-2014\footnote{\url{https://trec.nist.gov/data/webmain.html}} data, which was originally designed to evaluate search result diversification. The TREC collection contains 300 search topics.
The presence of varied search topics, which express different user information needs, helps us imitate open domain user-system interactions, which are required in~\ref{item:req_domain}.
Each topic in the collection is specific, ambiguous, or faceted~\cite{clarke2009overview}. In this work, we use the term `facet' to refer to the subtopics of both faceted and ambiguous topics. For clarity, the example of mapping from the search topic to set of facets is provided in Tab.~\ref{tab:facet_example}.
Faceted and ambiguous topics make an ideal case to study the effect of clarifying questions as they can be interpreted in various ways, which is required by~\ref{item:req_amb}. 
User information needs are expressed in the form of short search topic description\footnote{sometimes also can be referred as `keyword query'} (Tab.~\ref{tab:facet_example}) because the TREC Web track collection was designed for web search needs. Therefore, to satisfy requirement~\ref{item:req_conversational} and to make requests lexical diverse,
we ask expert annotators to convert those short keyword queries to the proper conversational request. The examples of such conversion are presented in Tab.~\ref{tab:query2con_req}. 
To satisfy the requirement~\ref{item:req_amb_score}, annotators are asked to provide a score for each request to reflect if clarification is needed ranging from 1 to 4, where `1' stands for very low or no need for clarification and `4' indicates a highly ambiguous request (examples are provided in Tab.~\ref{tab:query2con_req}). Two annotators assess clarification need of a query. In case of disagreement, we assign an additional annotator to make the final assessment. We achieve a high inter-annotator agreement on this task (Cohen's $\kappa = 0.78$).

\begin{table*}
    \caption{Examples of keyword queries converted to conversational requests with assigned clarification score (C. score).}
    \label{tab:query2con_req}
    \centering
    \begin{tabular}{llc}
    \toprule 
    \bf Search Topic & \bf Conversational Request         & C. Score\\ 
    \midrule
    average charitable donation & What is average charitable donation? & 1\\
    gmat prep classes           & How to prepare for the GMAT? & 2\\
    von willebrand disease      & What is von Willebrand Disease? &  3\\
    land surveyor               & I'm interested to know about land surveyor &  3\\
    alexian brothers hospital   & Give me information about Alexian Brothers hospitals & 3\\
    worm                        & I’m looking for information on worm & 4\\
    \bottomrule
    \end{tabular}
\end{table*}

\noindent
\mypar{\textbf{Collecting Clarifying Questions $\{Q\}$}} 
To collect $\{Q\}$ for every $U$ that satisfies~\ref{item:req_question}: (1)~we utilized Qulac dataset\footnote{\url{https://github.com/aliannejadi/qulac}} by converting topics into conversational requests;
(2)~we significantly extended it by crowdsourcing more data through Human Intelligence Task (HIT) on Amazon Mechanical Turk\footnote{\url{http://www.mturk.com}}, which design follows a general strategy proposed in~\cite{AliannejadiSigir19}.
Namely, we asked the workers to imagine themselves acting as a conversational agent\footnote{such as Microsoft Cortana, Alexa, or Google Assistant.} where an imaginary user had asked them about a topic. Then, we described the concept of facet to them, supporting it with multiple examples. Finally, we ask Turkers to do the following:
 \begin{itemize}[leftmargin=*,nosep]
    \item discover the facets of each $U$ using a preferred search engine and scan the results in the first three pages; and
    \item generate six questions related to $U$, aiming to address the facets they had figured out.
\end{itemize}

We assigned two workers per HIT, resulting in $12$ questions per $U$ in the first round. To preserve the questions' language diversity, we limited each worker to a maximum of two HITs. HITs were available to workers residing in the U.S. who had an approval rate of over 97\%.

\noindent
\mypar{\textbf{Controlling Quality of Clarifying Questions}}
To estimate the quality of the collected questions, we aim to address two main concerns:
(1)~\emph{how good are the collected clarifying questions?}; and 
(2)~\emph{Is the set of clarifying questions diverse (in other words, addressing different facets associated with the topic)?}
Given the high complexity of this task, we appointed two expert annotators. They were instructed to read all the collected questions on each topic, marking invalid and duplicate questions. Annotators were asked to match a question to a facet if its answer would address the facet. 
Finally, to ensure that all facets were covered by at least one question, we asked the annotators to generate an additional question for each facet that needed more specific questions. 

\noindent
\mypar{\textbf{Collecting Answers}}
To satisfy~\ref{item:req_answer}, we designed another HIT to collect coherent and consistent answers to the clarifying questions. The task started with detailed instructions followed by several examples. 
The workers were given $U$ and a facet description. Then we instruct them to assume that they had submitted the initial user request $U$ with their actual information need being the given facet.
Then workers were required to write the answer to the one clarifying question that was presented to them.
If a question required information other than what workers were provided with, they were instructed to use a `No answer' tag.
Each worker was allowed to complete a maximum of $100$ HITs to ensure language diversity. Workers were based in the U.S. with an approval rate of 95\% or greater.

\noindent
\mypar{\textbf{Controlling quality of Collected Answers}} During the course of data collection, we performed regular quality checks on the collected answers. The checks were done manually on 10\% of submissions per worker. In case we observed any invalid submissions among the sampled answers of one worker, we then studied all the submissions from that workers. Invalid submissions were then removed from the collection, and the worker was banned. Finally, we assigned all invalid answers to other workers to complete. Moreover, we employed basic behavioral check techniques in the design of the HIT. For example, we disabled copy/paste features of text inputs and tracked workers' keystrokes. This enabled us to detect and reject low-quality submissions.

As an outcome, we have a high-quality collection of single-turn conversations in the form of required triples: $(U,Q,A)$, which is marked as $P_1$ in our pipeline in Fig.~\ref{fig:pipeline}. Tab.~\ref{tab:stat_train} provides a statistics on collected dataset of single-turn conversations.

\subsection{P$_3$: Crowdsourcing Multi-Turn Dialogues}
\label{sec:multi}

The collected dataset of single-turn is sufficient to train and evaluate several conversational agents. More technical details on training and evaluation are provided in Sec.~\ref{sec:baselines}. For now, we assume that as a result of $P_2$: $\mathit{DA}$ is one of the best-performing trained dialogue agents.
We assume that the trained $\mathit{DA}$ can have a conversation with users. Namely, it should either ask a clarification question or give a factual answer to the user's request at each dialog step.
Therefore, the trained $\mathit{DA}$ is capable of:
\begin{itemize} [leftmargin=*,nosep]
    \item providing clarification question whenever appropriate in the course of the conversation;
    \item interpreting user's answer to the clarifying question.
\end{itemize}

\begin{figure}
    \centering
    \includegraphics[width=0.9\columnwidth]{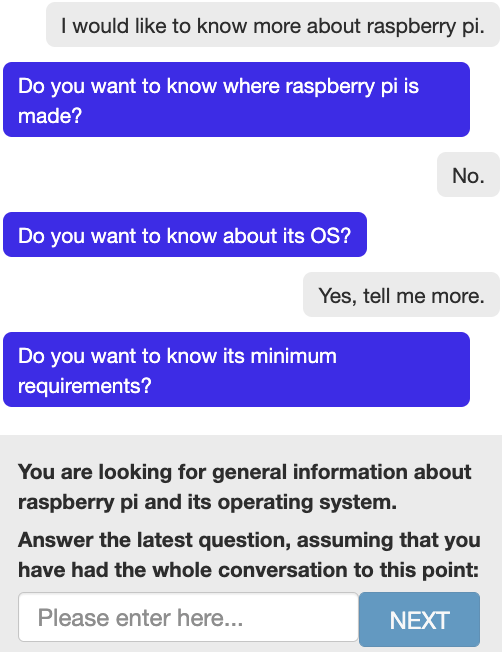}
    \caption{An example of the task provided at the HIT for multi-turn conversations, which asks to submit the answer given dialogue history/context.}
    \label{fig:multi_hit}
\end{figure}

To collect multi-turn conversations, we utilize best-performing dialogue agents that can accommodate an arbitrary number of turns, having two goals in mind:
\begin{enumerate} [leftmargin=*,label=\textbf{G\arabic*},nosep]
    \item \label{item:goal_evaluating} evaluating the quality of the agents in multi-turn settings where they converse with real humans; and 
    \item \label{item:goal_collecting} collecting a new dataset of multi-turn conversations with respect to clarification questions.  
\end{enumerate}

To reach~\ref{item:goal_evaluating}, the idea is to run the agents multiple times with different turns and evaluate them accordingly.
For that purpose, we design a HIT similar to the ones described in Sec.~\ref{sec:single} with the difference in the context of a conversation. We instructed crowd workers to understand the user's actual information need and imagine they are looking for the same information. Then, follow a conversation on that as presented in Fig.~\ref{fig:multi_hit}. The workers were instructed to answer the last question in the conversation while considering the conversation's context, which consists of previous questions and answers. The context of conversation could include 1-2 rounds of question-answer interactions.
To check the quality of clarifying questions returned by the trained dialogues, we instructed workers to indicate if the question was not understandable or was not in a proper language. As a result, such questions were removed from the collection.
We use the same quality check procedure for the collected answers as described in the previous section.
Tab.~\ref{tab:stat_train} provides a statistics on collected multi-turn conversations, to achieve \ref{item:goal_collecting}.

\mypar{\textbf{Synthetic Multi-Turn Conversations}} We also generate synthetic multi-turn conversations for training purposes. To do so, for each topic, we create a set of all possible combinations of questions (2 or 3 questions) together with their corresponding answers. 

\begin{table}[]
    \centering
    \caption{Statistics over collected data.}
    \label{tab:stat_train}
    \resizebox{1.0\columnwidth}{!}{
        \begin{tabular}{ll}
            \toprule
             \# search topics & 298 \\
            \# faceted \& ambiguous topics & 250\\
            \# single topics & 48\\
             \# facets & 1070\\
             \# questions & 3,304\\
             Average terms per question & 9.74 $\pm$ 2.62\\
             Average terms per answer & 8.48 $\pm$ 4.40\\
             \midrule 
             \# synthetic conversations (all) & 1,596,757\\
             \# synthetic conversations (2 turns)  & 203,050\\
             \# synthetic conversations (3 turns)  & 1,393,707\\
             \# human-machine dialogues (1 turn)  & 15,226\\
             \# human-machine dialogues (3 turns) & 499 \\
             \bottomrule
        \end{tabular}
        }
    \label{tab:stats_train}
\end{table}
\begin{table*}
    \caption{Performance for predictors returning classification need score based on \emph{dev} and \emph{test} sets. The best-performing model is marked in bold.}
    \centering
    \resizebox{0.6\textwidth}{!}{
    \begin{tabular}{lccccc}
    \toprule 
    Model  & & Precision & Recall & F1-Measure & MSE  \\  
        \midrule
        \cmidrule{2-6}
        \multirow{2}{*}{RoBERTa-based}
        & dev  & 0.6039 & 0.5600 & 0.5551 & 0.6200 \\
        & test & \bf 0.5981 & \bf 0.6557 & \bf 0.6070 & \bf 0.5409 \\
        \cmidrule{2-6}
        \multirow{2}{*}{BART}
        & dev   & 0.7008 & 0.7000 & 0.6976 & 0.5200 \\
        & test  & 0.4813 & 0.4754 & 0.4756 & 0.7705 \\  
        \cmidrule{2-6}
        \multirow{2}{*}{BERT-based}
        & dev  & 0.5218 & 0.4800 & 0.5000 & 0.8200\\
        & test & 0.3931 & 0.4918 & 0.4253 & 0.6557\\
        \bottomrule
    \end{tabular}
    }
\label{tab:clarification_need}    
\end{table*}

\section{Models and Evaluation}
\label{sec:baselines}

Following the suggested pipeline in Fig.~\ref{fig:pipeline}, we explain our contributions regarding the \emph{evaluation} of \emph{`asking clarifying questions in open-domain dialogues'} problem, namely:
\begin{itemize}[leftmargin=*,nosep]
    \item $P_2$: how the dialogue agents are trained and automatically evaluated based on single-turn conversations (Sec.~\ref{sec:eval_single_turn});
    \item $P_4$: how evaluation of multi-turn conversations is performed from both perspectives: offline automatic manner and having human-in-the-loop using Acute-eval~\cite{li2019acute} (Sec.~\ref{sec:acute_eval}).
\end{itemize}
\noindent
We design our experiments to collect answers to the following research questions:
\begin{enumerate}[leftmargin=*,label=\textbf{RQ\arabic*},nosep]
   \item \label{item:rq_when} When to ask clarifying questions during open-domain dialogues? (Sec.~\ref{sec:task1})
   \item \label{item:rq_which} Which clarifying question to ask for a given context of a conversation? (\textbf{a.} the single-turn conversations case is described in Sec.~\ref{sec:task2}; \textbf{b.} multi-turn one -- Sec.~\ref{sec:acute_eval})
\end{enumerate}

\subsection{P$_2$: Evaluating Single-Turn Agents}
\label{sec:eval_single_turn}

The collected dataset, described in Sec.~\ref{sec:single}, is split into training (70\%), validation (dev) (10\%), and test (20\%) sets. We split the data based on the search topic and maintained the same split for all single-turn and multi-turn experiments.
During the evaluation procedure, the following is used: (1)~ a set of conversational user requests, and (2)~a set of questions (i.e., question bank), which contains all collected questions on all the topics.

\subsubsection{Predicting Clarification Need}
\label{sec:task1}
\mypar{\textbf{Task}} The task is, given a user request, return a score from $1$ (no need for clarifying questions) to $4$ (cannot provide any answers without user clarification) indicating the necessity of asking clarifying questions (as depicted in module `Understanding User Request' in Fig.~\ref{fig:problem}).

\noindent
\mypar{\textbf{Automatic Evaluation}}
To evaluate the performance of the suggested classifier, we use Precision, Recall, F1-Measure, and Mean Squared Error (MSE).
Tab.~\ref{tab:clarification_need} presents the collected results of various classification methods, which includes Roberta-based classifier~\cite{LiuRoberta_2019}, BART~\cite{chipman2010bart}, and BERT-based classifier~\cite{devlin2018bert}.
Based on the supplied results, we can answer~\ref{item:rq_when}: the task is rather difficult and potentially can benefit from more exploration despite the reasonable performance of the proposed baselines.

\subsubsection{Returning Clarifying Question}
\label{sec:task2}

\mypar{\textbf{Task}} The task is, given a user request which needs clarification, return the most suitable clarifying question from the supplied question bank (as shown in module CQG in Fig.~\ref{fig:problem}).

\noindent
\mypar{\textbf{Automatic Evaluation}} We introduce two main strategies for evaluation: (1)~document relevance and (2)~question relevance. 

\noindent
\mypar{\emph{Document Relevance}}
To estimate the relevance of the retrieved documents we use the following standard metrics: Mean Reciprocal Rank (MRR)~\cite{voorhees1999proceedings,radev2002evaluating}, Precision (P)@[1,3,5,10,20], 
Normalized Discounted Cumulative Gain (nDCG)@[1,3,5,20]~\cite{wang2013theoretical}.
These metrics are computed as follows: a selected clarifying question, together with its corresponding answer, is added to the original user request.
The updated query is then used to retrieve (or re-rank) documents from the collection. The quality of the question is then evaluated by measuring  how much the question and its answer affect document retrieval performance when added to the initial request. We evaluate document relevance based on the relevance assessments provided by the TREC Web Track.

\begin{table}
    \caption{A set of document relevance related metrics reported on \emph{dev} and \emph{test} sets. NDCG@3 (in bold) reported on the test set is used as the main metric to rank the quality of the models.}
    \label{tab:document_relevance}
    \centering
    \resizebox{1.0\columnwidth}{!}{
    \begin{tabular}{lccccc}
    \toprule 
    \bf Model  & & \bf MRR & \bf P@1 & \bf NDCG@3 & \bf NDCG@5 \\  
        \midrule
        \multirow{2}{*}{B$_1$: Worst Q.}
        & dev  & 0.0841 & 0.0125 & 0.0252 & 0.0313 \\
        & test & 0.0541 & 0.0000 & \bf0.0097 & 0.0154\\
        \cmidrule{2-6}
        \multirow{2}{*}{B$_2$: No Q.}   
        & dev & 0.3000 & 0.2063 & 0.1475 & 0.1530 \\
        & test  & 0.3223 & 0.2268 & \bf0.1134 & 0.1059 \\ 
        \cmidrule{2-6}
        \multirow{2}{*}{B$_3$: Best Q.}
        & dev  & 0.4882 & 0.4187 & 0.3337 & 0.3064 \\
        & test & 0.4881 & 0.4275 & \bf0.2107 & 0.1759 \\
        \midrule
        \multirow{2}{*}{M$_1$: Roberta}
        & dev  & 0.3640 & 0.2813 & 0.2002 & 0.1954 \\
        & test & 0.3190 & 0.2342 & \bf0.1265 & 0.1130 \\  
        \cmidrule{2-6}
        \multirow{2}{*}{M$_2$: ELECTRA}
        & dev & 0.3761 & 0.3000 & 0.2113 & 0.1955 \\
        & test & 0.3140 & 0.2379 & \bf 0.1229 & 0.1097 \\
        \cmidrule{2-6}
        \multirow{2}{*}{M$_3$: BERT}
        & dev  & 0.3596 & 0.2750 & 0.1879 & 0.1882 \\
        & test & 0.3044 & 0.2119 & \bf0.1131 & 0.1021 \\
        \cmidrule{2-6}        
        \multirow{2}{*}{M$_4$: BM25}
        & dev  & 0.3096 & 0.2313 & 0.1608 & 0.1530 \\
        & test & 0.3134 & 0.2193 & \bf0.1151 & 0.1061 \\
        \cmidrule{2-6}
        \multirow{2}{*}{M$_5$: \makecell{BERT \\ +BM25}}
         & dev  & 0.3180 & 0.2437 & 0.1625 & 0.1550 \\
         & test & 0.3216 & 0.2453 & \bf0.1196 & 0.1097 \\
        \cmidrule{2-6} 
        \multirow{2}{*}{M$_6$: \makecell{Roberta \\ +BM25}}
        & dev  &  0.3606 & 0.2813 & 0.1942 & 0.1891 \\
        & test & 0.3045 & 0.2156 & \bf0.1108 & 0.1025 \\  
        \bottomrule
    \end{tabular}
    }
\end{table}

\noindent
\mypar{\emph{Question Relevance}}
Models are also evaluated in how well they can rank relevant questions higher than other questions in the question bank. For this task, which we call `question relevance,' the models are evaluated in terms of Recall@[10,20,30]. Since the precision of models is evaluated in the document relevance task, here we focus only on recall.

The suggested evaluation metrics are collected for a number of baselines (\textbf{B}) and fine-tuned state-of-the-art NLP models (\textbf{M}):
\begin{itemize}[leftmargin=*,nosep]
    \item \textbf{B$_1$: Worst Question}, when the dialogue system returns the least relevant question; 
    \item \textbf{B$_2$: No Question}, when the system never returns any clarifying question;
    \item \textbf{B$_3$: Best Question}, which show oracle performance as it always returns the most relevant question from the bank. 
    \item \textbf{M$_1$:} fine-tuned version of the pre-trained Roberta~\cite{LiuRoberta_2019}; 
    \item \textbf{M$_2$:} fine-tuned sequence classification model based on pre-trained ELECTRA~\cite{clark2020electra,ou2020clarifying};
    \item \textbf{M$_3$:} fine-tuned version of pre-trained BERT~\cite{devlin2018bert}
    \item \textbf{M$_4$:} BM25~\cite{robertson1995okapi}; and
    \item `+BM25' for \textbf{M$_5$} and \textbf{M$_6$:} means that BM25 on top on neural baseline for the final re-ranking of the questions from the bank.
\end{itemize}
Based on results of automatic evaluations of all the methods suggested above are reported in Tab.~\ref{tab:document_relevance} and Tab~\ref{tab:question_relevance} for single-turn conversations, to answer~\ref{item:rq_which}\textbf{.a}, we can conclude the performance of the best-performing fine-tuned neural SoTAs is reasonable and we can use them for multi-turn conversation.

\begin{table}
    \caption{A set of question relevance related metrics reported on \emph{dev} and \emph{test} sets. Recall@30 (in bold) reported on the test set is used as the main metric to rank the quality of the models.}
    \label{tab:question_relevance}
    \centering
    \resizebox{1.0\columnwidth}{!}{
    \begin{tabular}{lccccc}
    \toprule 
    Model  & &  R@5 & R@10 & R@20 & \bf R@30 \\  
        \midrule
        \multirow{2}{*}{M$_1$: Roberta}   
        & dev  & 0.3649 & 0.6694 & 0.8265 & 0.8587  \\ 
        & test & 0.3395 & 0.6251 & 0.8176 & \bf 0.8568  \\
        \cmidrule{2-6}
        \multirow{2}{*}{M$_2$: ELECTRA}
        & dev  & 0.3604 & 0.6749 & 0.8478 & 0.8761 \\
        & test & 0.3404 & 0.6329 & 0.8335 & \bf 0.8744 \\ 
        \cmidrule{2-6}
        \multirow{2}{*}{M$_3$: BERT}
        & dev  & 0.3492 & 0.6196 & 0.7337 & 0.7632 \\
        & test & 0.3438 & 0.6228 & 0.7987 & \bf 0.8409\\ 
        \cmidrule{2-6}
        \multirow{2}{*}{M$_4$: BM25}
        & dev   & 0.3245 & 0.5638 & 0.6675 & 0.6913  \\
        & test  & 0.3170 & 0.5705 & 0.7292 & \bf 0.7682 \\
        \cmidrule{2-6}         
        \multirow{2}{*}{M$_5$: \makecell{BERT \\ +BM25}}
        & dev  & 0.3454 & 0.6166 & 0.7354 & 0.7621 \\
        & test & 0.3272 & 0.6061 & 0.8013 & \bf 0.8433 \\
        \cmidrule{2-6}
        \multirow{2}{*}{M$_6$: \makecell{Roberta \\ +BM25}}
        & dev  & 0.3637 & 0.6409 & 0.7484 & 0.7793  \\
        & test & 0.3361 & 0.6219 & 0.7960 & \bf 0.8360  \\
        \bottomrule
    \end{tabular}
    }
\end{table}

\subsection{P$_4$: Evaluating Multi-Turn Conversations}
\label{sec:acute_eval}

\noindent
\mypar{\textbf{Task}} The task is, given an ongoing conversation with multiple turns, select or generate the next question that would clarify the user's intent best. The main goal is to learn from previous user feedback and ask a question that would lead to the highest information gain. 

\noindent
\mypar{\textbf{Automatic Evaluation}} Similar to the single-turn task, we evaluate the effectiveness of the baseline models based on document relevance. Therefore, we utilize the whole conversation context, clarifying questions, and human responses to retrieve documents from the collection and assess the quality of a question based on its impact on ranking performance. 
Note that we do not evaluate multi-turn models in terms of question relevance, since the question relevance is intended to evaluate recall of questions related to the search topic.
Due to complexity and costs of the evaluation, we pick two best-performing models from Sec.~\ref{sec:task2} for this task. To do so, we use the synthetic training data to fine-tune ELECTRA and Roberta similarly to our single-turn setup, but in the multi-turn case the whole history context is considered as a user request. The module for deciding whenever the request needs clarification is preserved.  
We see in Tab.~\ref{tab:multi_eval} that ELECTRA outperforms Roberta in terms of all evaluation metrics by a margin. One promising future research line might be exploring what properties of these two models lead to that difference in their effectiveness for this task.

\begin{table}
    \caption{A set of document relevance related metrics reported on test set for multi-turn conversations. NDCG@3 (in bold) is used as the main metric to rank the quality of the models.}
    \centering
    \resizebox{1.0\columnwidth}{!}{
    \begin{tabular}{lcccc}
    \toprule 
    \bf Model  & \bf MRR & \bf P@1 & \bf NDCG@3 & \bf NDCG@5 \\  
        \midrule
        ELECTRA & 0.1798 & 0.1161 &	\bf 0.0553 & 0.0536  \\
        Roberta & 0.1669 & 0.1067 & \bf 0.0522 & 0.0494  \\
        \bottomrule
    \end{tabular}
    }
\label{tab:multi_eval}    
\end{table}

\begin{table}
    \caption{Pairwise comparison of ELECTRA and Roberta on the multi-turn conversations. The values report the percentage of judgements in which ELECTRA wins Roberta in terms of  HU, EG, IN, KL, and CL.\footnote{Abbreviations are explained in the text.}}
    \centering
    \resizebox{1.0\columnwidth}{!}{
    \begin{tabular}{lccccc}
    \toprule 
    \bf Comparison  & \bf HU & \bf EG & \bf IN & \bf KL & \bf CL \\  
        \midrule
        ELECTRA vs.~Roberta & 0.57 & 0.59 &	0.56 & 0.57 & 0.56  \\
        \bottomrule
    \end{tabular}
    }
    \vspace{-0.5cm}
\label{tab:human_eval}    
\end{table}

\begin{figure}
    \centering
    \includegraphics[width=0.9\columnwidth]{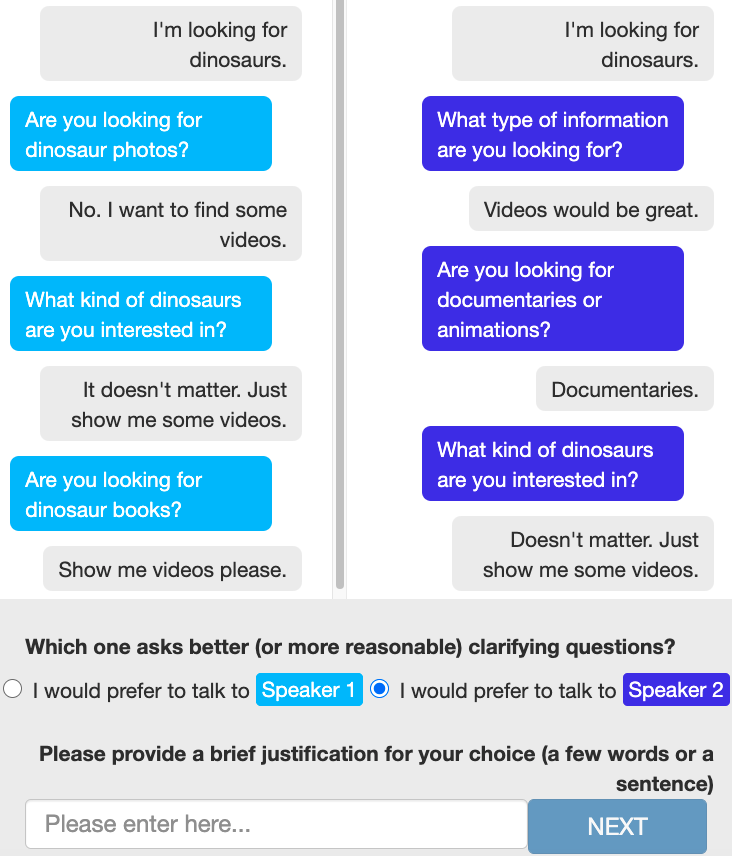}
    \caption{Sample on-boarding dialogue for clarification annotation using Acute-eval framework.}
    \label{fig:clarification_annotation_sample}
\end{figure}

\noindent
\mypar{\textbf{Human Evaluation}} To ensure that our automatic evaluation reflects the dialogues' quality, we conduct a pairwise human evaluation on two of the baselines. We use and extend the Acute-eval human annotation framework~\cite{li2019acute} to evaluate 120 randomly sampled dialogue pairs. For consistency, we use the four questions that the authors suggest measuring Humanness (HU), Engangingness (EG), Interestingness (IN), Knowledgeable (KL), and add a fifth one, specific to our task, on Clarification (CL). We modify the crowdsourcing task to inform the annotators about the conversation's main goal (i.e., information seeking). Furthermore, it is crucial to ensure that the annotators consider the model's ability to understand the user's feedback and incorporate the additional knowledge when asking its next question. Therefore, we added another question to examine this aspect of the conversation. As shown in Fig.~\ref{fig:clarification_annotation_sample}, after showing two full conversations to the annotators, they evaluated the model's ability of clarification by answering the following question: \emph{`Which one asks better (or more reasonable) clarifying questions?'}. Tab.~\ref{tab:human_eval} reports the results of our human evaluation of 120 dialogue pairs in terms of percentage of cases that ELECTRA beats Roberta based on the human annotation. We see that ELECTRA is judged to be the best model in most cases for all five aspects. It is interesting to see that the human annotation is in line with the proposed automatic annotation, suggesting that our approach approximates the true quality of the models. Moreover, we see that our new evaluation dimension, Clarification, achieves a similar result to other dimensions, which suggests that it should be included in the evaluation framework for the open-domain dialogues.  

Based on reported results for automatic evaluation (Tab.~\ref{tab:multi_eval}), which is aligned with human-in-a-loop one (Tab.~\ref{tab:human_eval}), we can conclude that suggested methods can be solid baselines for the follow-up research as an answer~\ref{item:rq_which}\textbf{.b}.

\section{Conclusions}
\label{sec:conclusion}

Asking clarifying questions when a user request is ambiguous is essential for developing human-like open-domain dialogue systems. 
In this work, we introduce a large-scale dataset that covers almost 300 different topics and is suitable as a foundation for further research. The collected dataset includes both single- and multi-turn conversations. 
We benchmark several state-of-the-art neural models that were fine-tuned for asking clarifying questions. Based on how these models performed, we conclude that they are solid baselines for future research that more fully explores the problem space.
In this paper, we also suggest an offline automatic evaluation pipeline, which agrees with human-in-loop evaluation.

We  publicly release the collected datasets, the code for training baselines, and our evaluation procedure in order to push forward the state-of-the-art.

\section*{Acknowledgement}
This work was supported in part by the NWO Innovational Research Incentives Scheme Vidi (016.Vidi.189.039) and an Engineering and Physical Sciences Research Council grant EP/V025708/1. We thank Microsoft Research, Google, and Amazon Science for their support to collect the data and organize the ConvAI3 data challenge. We also thank all the participants of the data challenge for taking part in the competition and releasing their models.
Any opinions, findings and conclusions or recommendations expressed in this material are those of the authors and do not necessarily reflect those of the sponsors. 

\bibliographystyle{acl_natbib}
\bibliography{acl2021,anthology}

\begin{thebibliography}{61}
\expandafter\ifx\csname natexlab\endcsname\relax\def\natexlab#1{#1}\fi

\bibitem[{Adiwardana et~al.(2020)Adiwardana, Luong, So, Hall, Fiedel,
  Thoppilan, Yang, Kulshreshtha, Nemade, Lu et~al.}]{adiwardana2020towards}
Daniel Adiwardana, Minh-Thang Luong, David~R So, Jamie Hall, Noah Fiedel, Romal
  Thoppilan, Zi~Yang, Apoorv Kulshreshtha, Gaurav Nemade, Yifeng Lu, et~al.
  2020.
\newblock Towards a human-like open-domain chatbot.
\newblock \emph{arXiv preprint arXiv:2001.09977}.

\bibitem[{Aliannejadi et~al.(2021)Aliannejadi, Azzopardi, Zamani, Thomas, and
  Craswell}]{aliannejadi2021cikm}
Mohammad Aliannejadi, Leif Azzopardi, Hamed Zamani, Paul Thomas, and Nick
  Craswell. 2021.
\newblock Analysing mixed initiatives and search strategies during
  conversational search.
\newblock In \emph{International Conference on Information and Knowledge
  Management ({CIKM})}.

\bibitem[{Aliannejadi et~al.(2020)Aliannejadi, Kiseleva, Chuklin, Dalton, and
  Burtsev}]{DBLP:journals/corr/abs-2009-11352}
Mohammad Aliannejadi, Julia Kiseleva, Aleksandr Chuklin, Jeff Dalton, and
  Mikhail~S. Burtsev. 2020.
\newblock {ConvAI3}: Generating clarifying questions for open-domain dialogue
  systems ({ClariQ}).
\newblock \emph{arXiv preprint arXiv:2009.11352}.

\bibitem[{Aliannejadi et~al.(2019)Aliannejadi, Zamani, Crestani, and
  Croft}]{AliannejadiSigir19}
Mohammad Aliannejadi, Hamed Zamani, Fabio Crestani, and W.~Bruce Croft. 2019.
\newblock Asking clarifying questions in open-domain information-seeking
  conversations.
\newblock In \emph{{ACM} {SIGIR} Conference on Research and Development in
  Information Retrieval (SIGIR)}, pages 475--484.

\bibitem[{Braslavski et~al.(2017)Braslavski, Savenkov, Agichtein, and
  Dubatovka}]{braslavski2017you}
Pavel Braslavski, Denis Savenkov, Eugene Agichtein, and Alina Dubatovka. 2017.
\newblock What do you mean exactly? analyzing clarification questions in {CQA}.
\newblock In \emph{Conference on Conference Human Information Interaction and
  Retrieval (CHIIR)}, pages 345--348.

\bibitem[{Burtsev et~al.(2017)Burtsev, Chuklin, Kiseleva, and
  Borisov}]{burtsev2017search}
Mikhail Burtsev, Aleksandr Chuklin, Julia Kiseleva, and Alexey Borisov. 2017.
\newblock Search-oriented conversational {AI} ({SCAI}).
\newblock In \emph{ACM SIGIR International Conference on Theory of Information
  Retrieval (ICTIR)}, pages 333--334.

\bibitem[{Burtsev and Logacheva(2020)}]{burtsev2020conversational}
Mikhail Burtsev and Varvara Logacheva. 2020.
\newblock Conversational intelligence challenge: Accelerating research with
  crowd science and open source.
\newblock \emph{AI Magazine}, 41(3):18--27.

\bibitem[{Burtsev et~al.(2018)Burtsev, Logacheva, Malykh, Serban, Lowe,
  Prabhumoye, Black, Rudnicky, and Bengio}]{burtsev2018first}
Mikhail Burtsev, Varvara Logacheva, Valentin Malykh, Iulian~Vlad Serban, Ryan
  Lowe, Shrimai Prabhumoye, Alan~W Black, Alexander Rudnicky, and Yoshua
  Bengio. 2018.
\newblock The first conversational intelligence challenge.
\newblock In \emph{The NIPS'17 Competition: Building Intelligent Systems},
  pages 25--46.

\bibitem[{Chai and Wan(2020)}]{chai-wan-2020-learning}
Zi~Chai and Xiaojun Wan. 2020.
\newblock Learning to ask more: Semi-autoregressive sequential question
  generation under dual-graph interaction.
\newblock In \emph{Annual Meeting of the Association for Computational
  Linguistics (ACL)}, pages 225--237.

\bibitem[{Chipman et~al.(2010)Chipman, George, McCulloch
  et~al.}]{chipman2010bart}
Hugh~A Chipman, Edward~I George, Robert~E McCulloch, et~al. 2010.
\newblock Bart: Bayesian additive regression trees.
\newblock \emph{The Annals of Applied Statistics}, 4(1):266--298.

\bibitem[{Clark et~al.(2020)Clark, Luong, Le, and Manning}]{clark2020electra}
Kevin Clark, Minh{-}Thang Luong, Quoc~V. Le, and Christopher~D. Manning. 2020.
\newblock {ELECTRA:} pre-training text encoders as discriminators rather than
  generators.
\newblock In \emph{International Conference on Learning Representations
  ({ICLR})}.

\bibitem[{Clarke et~al.(2009)Clarke, Craswell, and
  Soboroff}]{clarke2009overview}
Charles L.~A. Clarke, Nick Craswell, and Ian Soboroff. 2009.
\newblock Overview of the {TREC} 2009 web track.
\newblock In \emph{Text REtrieval Conference ({TREC})}.

\bibitem[{Dalton et~al.(2020)Dalton, Chuklin, Kiseleva, and
  Burtsev}]{dalton2020proceedings}
Jeff Dalton, Aleksandr Chuklin, Julia Kiseleva, and Mikhail Burtsev. 2020.
\newblock Proceedings of the 5th international workshop on search-oriented
  conversational {AI} ({SCAI}).
\newblock In \emph{International Workshop on Search-Oriented Conversational AI
  (SCAI)}.

\bibitem[{Devlin et~al.(2019)Devlin, Chang, Lee, and
  Toutanova}]{devlin2018bert}
Jacob Devlin, Ming{-}Wei Chang, Kenton Lee, and Kristina Toutanova. 2019.
\newblock {BERT:} pre-training of deep bidirectional transformers for language
  understanding.
\newblock In \emph{Conference of the North American Chapter of the Association
  for Computational Linguistics: Human Language Technologies ({NAACL-HLT})},
  pages 4171--4186.

\bibitem[{Dinan et~al.(2020)Dinan, Logacheva, Malykh, Miller, Shuster, Urbanek,
  Kiela, Szlam, Serban, Lowe et~al.}]{dinan2020second}
Emily Dinan, Varvara Logacheva, Valentin Malykh, Alexander Miller, Kurt
  Shuster, Jack Urbanek, Douwe Kiela, Arthur Szlam, Iulian Serban, Ryan Lowe,
  et~al. 2020.
\newblock The second conversational intelligence challenge ({ConvAI}2).
\newblock In \emph{The NeurIPS'18 Competition}, pages 187--208.

\bibitem[{Gao et~al.(2019)Gao, Li, King, and
  Lyu}]{gao-etal-2019-interconnected}
Yifan Gao, Piji Li, Irwin King, and Michael~R. Lyu. 2019.
\newblock Interconnected question generation with coreference alignment and
  conversation flow modeling.
\newblock In \emph{Annual Meeting of the Association for Computational
  Linguistics (ACL)}, pages 4853--4862.

\bibitem[{Hancock et~al.(2019)Hancock, Bordes, Mazar\'{e}, and
  Weston}]{hancock2019feed}
Braden Hancock, Antoine Bordes, Pierre-Emmanuel Mazar\'{e}, and Jason Weston.
  2019.
\newblock Learning from dialogue after deployment: Feed yourself, chatbot!
\newblock In \emph{Annual Meeting of the Association for Computational
  Linguistics (ACL)}, pages 3667--3684.

\bibitem[{Huang et~al.(2020)Huang, Zhu, and Gao}]{huang2020challenges}
Minlie Huang, Xiaoyan Zhu, and Jianfeng Gao. 2020.
\newblock Challenges in building intelligent open-domain dialog systems.
\newblock \emph{ACM Transactions on Information Systems (TOIS)}, 38(3):1--32.

\bibitem[{Jiang et~al.(2015)Jiang, Leung, Yang, and Ng}]{jiang2015query}
Di~Jiang, Kenneth Wai-Ting Leung, Lingxiao Yang, and Wilfred Ng. 2015.
\newblock Query suggestion with diversification and personalization.
\newblock \emph{Knowledge-Based Systems}, 89:553--568.

\bibitem[{Kato and Tanaka(2016)}]{kato2016suggest}
Makoto~P Kato and Katsumi Tanaka. 2016.
\newblock To suggest, or not to suggest for queries with diverse intents:
  Optimizing search result presentation.
\newblock In \emph{ACM International Conference on Web Search and Data Mining
  (WSDM)}, pages 133--142.

\bibitem[{Kiseleva et~al.(2016{\natexlab{a}})Kiseleva, Williams,
  Hassan~Awadallah, Crook, Zitouni, and Anastasakos}]{kiseleva2016predicting}
Julia Kiseleva, Kyle Williams, Ahmed Hassan~Awadallah, Aidan~C Crook, Imed
  Zitouni, and Tasos Anastasakos. 2016{\natexlab{a}}.
\newblock Predicting user satisfaction with intelligent assistants.
\newblock In \emph{ACM SIGIR Conference on Research and Development in
  Information Retrieval (SIGIR)}, pages 45--54.

\bibitem[{Kiseleva et~al.(2016{\natexlab{b}})Kiseleva, Williams, Jiang,
  Hassan~Awadallah, Crook, Zitouni, and
  Anastasakos}]{kiseleva2016understanding}
Julia Kiseleva, Kyle Williams, Jiepu Jiang, Ahmed Hassan~Awadallah, Aidan~C
  Crook, Imed Zitouni, and Tasos Anastasakos. 2016{\natexlab{b}}.
\newblock Understanding user satisfaction with intelligent assistants.
\newblock In \emph{Conference on Human Information Interaction and Retrieval
  (CHIIR)}, pages 121--130.

\bibitem[{Kocielnik et~al.(2019)Kocielnik, Amershi, and
  Bennett}]{kocielnik2019will}
Rafal Kocielnik, Saleema Amershi, and Paul~N Bennett. 2019.
\newblock Will you accept an imperfect {AI}? exploring designs for adjusting
  end-user expectations of {AI} systems.
\newblock In \emph{CHI Conference on Human Factors in Computing Systems (CHI)},
  pages 1--14.

\bibitem[{Kong and Allan(2014)}]{kong2014extending}
Weize Kong and James Allan. 2014.
\newblock Extending faceted search to the general web.
\newblock In \emph{ACM International Conference on Conference on Information
  and Knowledge Management (CIKM)}, pages 839--848.

\bibitem[{Kong and Allan(2016)}]{kong2016precision}
Weize Kong and James Allan. 2016.
\newblock Precision-oriented query facet extraction.
\newblock In \emph{ACM International on Conference on Information and Knowledge
  Management (CIKM)}, pages 1433--1442.

\bibitem[{Krasakis et~al.(2020)Krasakis, Aliannejadi, Voskarides, and
  Kanoulas}]{DBLP:conf/ictir/KrasakisAVK20}
Antonios~Minas Krasakis, Mohammad Aliannejadi, Nikos Voskarides, and Evangelos
  Kanoulas. 2020.
\newblock Analysing the effect of clarifying questions on document ranking in
  conversational search.
\newblock In \emph{{ACM} {SIGIR} International Conference on the Theory of
  Information Retrieval (ICTIR)}, pages 129--132.

\bibitem[{Kwiatkowski et~al.(2019)Kwiatkowski, Palomaki, Redfield, Collins,
  Parikh, Alberti, Epstein, Polosukhin, Devlin, Lee
  et~al.}]{kwiatkowski2019natural}
Tom Kwiatkowski, Jennimaria Palomaki, Olivia Redfield, Michael Collins, Ankur
  Parikh, Chris Alberti, Danielle Epstein, Illia Polosukhin, Jacob Devlin,
  Kenton Lee, et~al. 2019.
\newblock Natural questions: a benchmark for question answering research.
\newblock \emph{Transactions of the Association for Computational Linguistics},
  7:453--466.

\bibitem[{Li et~al.(2017)Li, Miller, Chopra, Ranzato, and
  Weston}]{li2016dialogue}
Jiwei Li, Alexander~H. Miller, Sumit Chopra, Marc'Aurelio Ranzato, and Jason
  Weston. 2017.
\newblock Dialogue learning with human-in-the-loop.
\newblock In \emph{International Conference on Learning Representations
  ({ICLR})}.

\bibitem[{Li et~al.(2019{\natexlab{a}})Li, Weston, and Roller}]{li2019acute}
Margaret Li, Jason Weston, and Stephen Roller. 2019{\natexlab{a}}.
\newblock Acute-eval: Improved dialogue evaluation with optimized questions and
  multi-turn comparisons.
\newblock \emph{arXiv preprint arXiv:1909.03087}.

\bibitem[{Li et~al.(2016)Li, Lipton, Dhingra, Li, Gao, and Chen}]{li2016user}
Xiujun Li, Zachary~C Lipton, Bhuwan Dhingra, Lihong Li, Jianfeng Gao, and
  Yun-Nung Chen. 2016.
\newblock A user simulator for task-completion dialogues.
\newblock \emph{arXiv preprint arXiv:1612.05688}.

\bibitem[{Li et~al.(2019{\natexlab{b}})Li, Kiseleva, and
  de~Rijke}]{li2018dialogue}
Ziming Li, Julia Kiseleva, and Maarten de~Rijke. 2019{\natexlab{b}}.
\newblock Dialogue generation: From imitation learning to inverse reinforcement
  learning.
\newblock In \emph{AAAI Conference on Artificial Intelligence (AAAI)}, pages
  6722--6729.

\bibitem[{Li et~al.(2020)Li, Lee, Peng, Li, Kiseleva, de~Rijke, Shayandeh, and
  Gao}]{li2020guided}
Ziming Li, Sungjin Lee, Baolin Peng, Jinchao Li, Julia Kiseleva, Maarten
  de~Rijke, Shahin Shayandeh, and Jianfeng Gao. 2020.
\newblock Guided dialogue policy learning without adversarial learning in the
  loop.
\newblock In \emph{Findings of the Association for Computational Linguistics
  ({EMNLP})}, pages 2308--2317.

\bibitem[{Li et~al.(2021)Li, Park, Kiseleva, Kim, and Lee}]{li2021data}
Ziming Li, Dookun Park, Julia Kiseleva, Young-Bum Kim, and Sungjin Lee. 2021.
\newblock {DEUS}: A data-driven approach to estimate user satisfaction in
  multi-turn dialogues.
\newblock \emph{arXiv preprint arXiv:2103.01287}.

\bibitem[{Liu et~al.(2019)Liu, Ott, Goyal, Du, Joshi, Chen, Levy, Lewis,
  Zettlemoyer, and Stoyanov}]{LiuRoberta_2019}
Yinhan Liu, Myle Ott, Naman Goyal, Jingfei Du, Mandar Joshi, Danqi Chen, Omer
  Levy, Mike Lewis, Luke Zettlemoyer, and Veselin Stoyanov. 2019.
\newblock {RoBERTa}: {A} robustly optimized {BERT} pretraining approach.
\newblock \emph{arXiv preprint arXiv:1907.11692}.

\bibitem[{Lotze et~al.(2021)Lotze, Klut, Aliannejadi, and
  Kanoulas}]{DBLP:journals/corr/abs-2103-06192}
Tom Lotze, Stefan Klut, Mohammad Aliannejadi, and Evangelos Kanoulas. 2021.
\newblock Ranking clarifying questions based on predicted user engagement.
\newblock \emph{arXiv preprint arXiv:2103.06192}.

\bibitem[{Lowe et~al.(2017)Lowe, Noseworthy, Serban, Angelard-Gontier, Bengio,
  and Pineau}]{lowe2017towards}
Ryan Lowe, Michael Noseworthy, Iulian~Vlad Serban, Nicolas Angelard-Gontier,
  Yoshua Bengio, and Joelle Pineau. 2017.
\newblock Towards an automatic turing test: Learning to evaluate dialogue
  responses.
\newblock In \emph{Annual Meeting of the Association for Computational
  Linguistics (ACL)}, pages 1116--1126.

\bibitem[{Nass and Moon(2000)}]{nass2000machines}
Clifford Nass and Youngme Moon. 2000.
\newblock Machines and mindlessness: Social responses to computers.
\newblock \emph{Journal of social issues}, 56(1):81--103.

\bibitem[{Novikova et~al.(2017)Novikova, Du{\v{s}}ek, Curry, and
  Rieser}]{novikova2017we}
Jekaterina Novikova, Ond{\v{r}}ej Du{\v{s}}ek, Amanda~Cercas Curry, and Verena
  Rieser. 2017.
\newblock Why we need new evaluation metrics for {NLG}.
\newblock In \emph{Conference on Empirical Methods in Natural Language
  Processing (EMNLP)}, pages 2241--2252.

\bibitem[{Ou and Lin(2020)}]{ou2020clarifying}
Wenjie Ou and Yue Lin. 2020.
\newblock A clarifying question selection system from ntes\_along in {ConvAI3}
  challenge.
\newblock \emph{arXiv preprint arXiv:2010.14202}.

\bibitem[{Radev et~al.(2002)Radev, Qi, Wu, and Fan}]{radev2002evaluating}
Dragomir~R. Radev, Hong Qi, Harris Wu, and Weiguo Fan. 2002.
\newblock Evaluating web-based question answering systems.
\newblock In \emph{International Conference on Language Resources and
  Evaluation ({LREC})}, pages 1153--1156.

\bibitem[{Radlinski and Dumais(2006)}]{radlinski2006improving}
Filip Radlinski and Susan Dumais. 2006.
\newblock Improving personalized web search using result diversification.
\newblock In \emph{{ACM} {SIGIR} Conference on Research and Development in
  Information Retrieval ({SIGIR})}, pages 691--692.

\bibitem[{Rao and III(2018)}]{rao2018learning}
Sudha Rao and Hal~Daum{\'{e}} III. 2018.
\newblock Learning to ask good questions: Ranking clarification questions using
  neural expected value of perfect information.
\newblock In \emph{Annual Meeting of the Association for Computational
  Linguistics (ACL)}, pages 2737--2746.

\bibitem[{Reiter(2018)}]{reiter2018structured}
Ehud Reiter. 2018.
\newblock A structured review of the validity of bleu.
\newblock \emph{Computational Linguistics}, 44(3):393--401.

\bibitem[{Robertson et~al.(1995)Robertson, Walker, Jones, Hancock-Beaulieu,
  Gatford et~al.}]{robertson1995okapi}
Stephen~E Robertson, Steve Walker, Susan Jones, Micheline~M Hancock-Beaulieu,
  Mike Gatford, et~al. 1995.
\newblock Okapi at {TREC}-3.
\newblock \emph{Nist Special Publication Sp}, 109:109.

\bibitem[{Roller et~al.(2021)Roller, Dinan, Goyal, Ju, Williamson, Liu, Xu,
  Ott, Smith, Boureau, and Weston}]{roller2020recipes}
Stephen Roller, Emily Dinan, Naman Goyal, Da~Ju, Mary Williamson, Yinhan Liu,
  Jing Xu, Myle Ott, Eric~Michael Smith, Y{-}Lan Boureau, and Jason Weston.
  2021.
\newblock Recipes for building an open-domain chatbot.
\newblock In \emph{Conference of the European Chapter of the Association for
  Computational Linguistics ({EACL})}, pages 300--325.

\bibitem[{Rosset et~al.(2020)Rosset, Xiong, Song, Campos, Craswell, Tiwary, and
  Bennett}]{rosset2020leading}
Corby Rosset, Chenyan Xiong, Xia Song, Daniel Campos, Nick Craswell, Saurabh
  Tiwary, and Paul Bennett. 2020.
\newblock Leading conversational search by suggesting useful questions.
\newblock In \emph{The Web Conference}, pages 1160--1170.

\bibitem[{See et~al.(2019{\natexlab{a}})See, Pappu, Saxena, Yerukola, and
  Manning}]{see2019massively}
Abigail See, Aneesh Pappu, Rohun Saxena, Akhila Yerukola, and Christopher~D.
  Manning. 2019{\natexlab{a}}.
\newblock Do massively pretrained language models make better storytellers?
\newblock In \emph{Conference on Computational Natural Language Learning
  (CoNLL)}, pages 843--861.

\bibitem[{See et~al.(2019{\natexlab{b}})See, Roller, Kiela, and
  Weston}]{see2019makes}
Abigail See, Stephen Roller, Douwe Kiela, and Jason Weston. 2019{\natexlab{b}}.
\newblock What makes a good conversation? how controllable attributes affect
  human judgments.
\newblock In \emph{Conference of the North American Chapter of the Association
  for Computational Linguistics: Human Language Technologies ({NAACL-HLT})},
  pages 1702--1723.

\bibitem[{Sekulic et~al.(2021)Sekulic, Aliannejadi, and
  Crestani}]{DBLP:conf/ecir/SekulicAC21}
Ivan Sekulic, Mohammad Aliannejadi, and Fabio Crestani. 2021.
\newblock User engagement prediction for clarification in search.
\newblock In \emph{European Conference on Information Retrieval ({ECIR})},
  pages 619--633.

\bibitem[{Soleimani et~al.(2021)Soleimani, Monz, and
  Worring}]{DBLP:conf/eacl/SoleimaniMW21}
Amir Soleimani, Christof Monz, and Marcel Worring. 2021.
\newblock {NLQuAD}: {A} non-factoid long question answering data set.
\newblock In \emph{Conference of the European Chapter of the Association for
  Computational Linguistics (EACL)}, pages 1245--1255.

\bibitem[{Su et~al.(2018)Su, Li, Gao, Liu, and Chen}]{Su2018D3Q}
Shang{-}Yu Su, Xiujun Li, Jianfeng Gao, Jingjing Liu, and Yun{-}Nung Chen.
  2018.
\newblock Discriminative deep dyna-q: Robust planning for dialogue policy
  learning.
\newblock In \emph{Conference on Empirical Methods in Natural Language
  Processing (EMNLP)}, pages 3813--3823.

\bibitem[{Takanobu et~al.(2019)Takanobu, Zhu, and Huang}]{takanobu2019guided}
Ryuichi Takanobu, Hanlin Zhu, and Minlie Huang. 2019.
\newblock Guided dialog policy learning: Reward estimation for multi-domain
  task-oriented dialog.
\newblock In \emph{Conference on Empirical Methods in Natural Language
  Processing and the International Joint Conference on Natural Language
  Processing (EMNLP-IJCNLP)}, pages 100--110.

\bibitem[{Trienes and Balog(2019)}]{trienes2019identifying}
Jan Trienes and Krisztian Balog. 2019.
\newblock Identifying unclear questions in community question answering
  websites.
\newblock In \emph{European Conference on Information Retrieval (ECIR)}, pages
  276--289.

\bibitem[{Vallet and Castells(2012)}]{vallet2012personalized}
David Vallet and Pablo Castells. 2012.
\newblock Personalized diversification of search results.
\newblock In \emph{ACM SIGIR conference on Research and Development in
  Information Retrieval (SIGIR)}, pages 841--850.

\bibitem[{Voorhees(1999)}]{voorhees1999proceedings}
EM~Voorhees. 1999.
\newblock Proceedings of the 8th text retrieval conference.
\newblock \emph{TREC-8 Question Answering Track Report}, pages 77--82.

\bibitem[{Wang et~al.(2018)Wang, Liu, Huang, and Nie}]{wang2018learning}
Yansen Wang, Chenyi Liu, Minlie Huang, and Liqiang Nie. 2018.
\newblock Learning to ask questions in open-domain conversational systems with
  typed decoders.
\newblock In \emph{Annual Meeting of the Association for Computational
  Linguistics (ACL)}, pages 2193--2203.

\bibitem[{Wang et~al.(2013)Wang, Wang, Li, He, and Liu}]{wang2013theoretical}
Yining Wang, Liwei Wang, Yuanzhi Li, Di~He, and Tie-Yan Liu. 2013.
\newblock A theoretical analysis of {NDCG} type ranking measures.
\newblock In \emph{Conference on Learning Theory (COLT)}, pages 25--54.

\bibitem[{Zamani et~al.(2020{\natexlab{a}})Zamani, Dumais, Craswell, Bennett,
  and Lueck}]{zamani2020generating}
Hamed Zamani, Susan Dumais, Nick Craswell, Paul Bennett, and Gord Lueck.
  2020{\natexlab{a}}.
\newblock Generating clarifying questions for information retrieval.
\newblock In \emph{The Web Conference}, pages 418--428.

\bibitem[{Zamani et~al.(2020{\natexlab{b}})Zamani, Lueck, Chen, Quispe, Luu,
  and Craswell}]{zamani2020mimics}
Hamed Zamani, Gord Lueck, Everest Chen, Rodolfo Quispe, Flint Luu, and Nick
  Craswell. 2020{\natexlab{b}}.
\newblock {MIMICS:} {A} large-scale data collection for search clarification.
\newblock In \emph{{ACM} International Conference on Information and Knowledge
  Management (CIKM)}, pages 3189--3196.

\bibitem[{Zhang et~al.(2018)Zhang, Dinan, Urbanek, Szlam, Kiela, and
  Weston}]{zhang2018personalizing}
Saizheng Zhang, Emily Dinan, Jack Urbanek, Arthur Szlam, Douwe Kiela, and Jason
  Weston. 2018.
\newblock Personalizing dialogue agents: I have a dog, do you have pets too?
\newblock In \emph{Annual Meeting of the Association for Computational
  Linguistics (ACL)}, pages 2204--2213.

\bibitem[{Zhang et~al.(2020)Zhang, Sun, Galley, Chen, Brockett, Gao, Gao, Liu,
  and Dolan}]{zhang2019dialogpt}
Yizhe Zhang, Siqi Sun, Michel Galley, Yen{-}Chun Chen, Chris Brockett, Xiang
  Gao, Jianfeng Gao, Jingjing Liu, and Bill Dolan. 2020.
\newblock {DIALOGPT} : Large-scale generative pre-training for conversational
  response generation.
\newblock In \emph{Annual Meeting of the Association for Computational
  Linguistics (ACL)}, pages 270--278.

\end{thebibliography}

\end{document}